# FROST FILTERED SCALE-INVARIANT FEATURE EXTRACTION AND MULTILAYER PERCEPTRON FOR HYPERSPECTRAL IMAGE CLASSIFICATION

G.Kalaiarasi[1], S.Maheswari[2]

[1] Research Scholar / ECE, Jai Shriram Engineering College, Tirupur, India

[2] Assistant Professor (Sr Gr) / EEE, Kongu Engineering College, Erode, India

**Abstract**

Hyperspectral image (HSI) classification plays a significant in the field of remote sensing due to its ability to provide spatial and spectral information. Due to the rapid development and increasing of hyperspectral remote sensing technology, many methods have been developed for HSI classification but still a lack of achieving the better performance. A Frost Filtered Scale-Invariant Feature Transformation based MultiLayer Perceptron Classification (FFSIFT-MLPC) technique is introduced for classifying the hyperspectral image with higher accuracy and minimum time consumption. The FFSIFT-MLPC technique performs three major processes, namely preprocessing, feature extraction and classification using multiple layers. Initially, the hyperspectral image is divided into number of spectral bands. These bands are given as input in the input layer of perceptron. Then the Frost filter is used in FFSIFT-MLPC technique for preprocessing the input bands which helps to remove the noise from hyper-spectral image at the first hidden layer. After preprocessing task, texture, color and object features of hyper-spectral image are extracted at second hidden layer using Gaussian distributive scale-invariant feature transform. At the third hidden layer, Euclidean distance is measured between the extracted features and testing features. Finally, feature matching is carried out at the output layer for hyper-spectral image classification. The classified outputs are resulted in terms of spectral bands (i.e., different colors). Experimental analysis is performed with PSNR, classification accuracy, false positive rate and classification time with number of spectral bands. The results evident that presented FFSIFT-MLPC technique improves the hyperspectral image classification accuracy, PSNR and minimizes false positive rate as well as classification time than the state-of-the-art methods.

*Keywords:* Hyperspectral image classification, Frost Filter based preprocessing, Gaussian distributive Scale-invariant feature transform, Multilayer Perceptron, Euclidean distance

## 1. Introduction

Remotely-sensed hyperspectral images (HSI) are images gathered from the satellites that report a wide range of the electromagnetic spectrum, generally more than hundred spectral bands from visible to near-infrared wavelengths. Typically, HSI is used in several applications, such as agriculture, disaster relief, military and so on. One of the most significant problems in HSI classification using hundreds of narrowband spectral bands provides the spectral information to identify the object.

Hybridized composite kernel boosting with extreme learning machines (HCKBoost) technique was developed in [1] for performing the classification task with the hyperspectral images. The designed technique was not improving the image quality and failed to extract more spatial features for achieving higher classification accuracy. A Guided Filter Support Vector Machine Edge Preserving Filter (GF-SVM-EPF) technique was introduced in [2] for improving the classification results of hyperspectral images. But the technique was not improved the classification accuracy and failed to minimize the time complexity.

A bilayer elastic net (ELN2) regression model was introduced in [3] for classifying the hyperspectral image using spectral-spatial information. The regression algorithm consumed more computation time. A fuzziness-based active learning method was introduced in [4] to enhance the performance of hyperspectral image categorization. The designed framework was not minimized the false positive rate of the classification.

A Convolutional Neural network-based classification technique was introduced in [5] for HSI classification using spectral-spatial features. The classification technique failed to focus on filtering parameters for improving the Peak Signal to Noise Ratio (PSNR). A novel Semisupervised-Entropy based approach was developed in [6] for categorizing the Hyperspectral Image using Random Forest. The designed approach consumed more running time for classifying the images.

A long short term memory (LSTM) networks were developed in [7] for hyperspectral image classification using spectral and spatial information concurrently. However, the performance of classification time remained unaddressed. The subpixel target detection technique was introduced in [8] for classifying the HIS image with higher precision ad minimum false classification. The technique failed to uses the other pixel-based spectral filters for removing the noise.

A Multiscale superpixel-level based support vector machine (MSP-SVM) was developed in [9] to present the classification using HIS images. But the false positive rate was not minimized using MSP-SVM. In [10], a Convolutional neural network (CNN) was developed for HSI classification using better feature representation. The classification time was not minimized. A Recurrent Neural Network (RNN) was developed in [11] to increase the classification performance of Hyperspectral Image. Though the RNN uses the filtering technique to remove the noise, the feature extraction was not performed using an efficient algorithm.

An iterative target-constrained interference-minimization classifier was developed in [12] to increase the performance of classification for multiple classes using the hyperspectral image. Though the classifier minimizes the misclassification rate, the classification time was not minimized. A multiple kernel learning method was introduced in [13] for hyperspectral image classification using a model of information entropy. But the performance of false-positive rate was not minimized.

An improved Rotation Forest (ROF) technique was developed in [14] for accurately performs the classification through the feature extraction. The method failed to filter the noise for achieving the higher accuracy. A joint sparse representation classification method was introduced in [15] using the hyperspectral image by extracting the features. However, the method failed to use discriminative learning algorithms to improve classification accuracy.

In [16], a Discriminative low-rank Gabor filtering (DLRGF) technique was developed to categorize the image using spectral-spatial information. Though the technique improves the classification accuracy and minimizes the computation time, the performance of other metrics such as false positive rate, noise removal remained unaddressed. A deep multiple feature fusion (DMFF) approach was introduced in [17] for categorizing the hyperspectral image. The designed approach failed to minimize the classification time.

A multi-grained network (MugNet) was introduced in [18] for categorizing the hyperspectral image. But the computational efficiency was not increased. A probabilistic support vector machine was developed in [19] using a Markov random field for image classification with the multiple features. The image quality was not improved to further achieving the higher classification accuracy. A joint spatial-spectral hyperspectral image classification technique was introduced in [20] using the two-stream convolutional network through the spatial-spectral feature. Though, it attains higher accuracy of classification, the time was not reduced.

### 1.1 Contribution of the paper

The major issues of conventional methods are higher complexity, lack of noise removal, minimum classification accuracy, higher false-positive rate, more time consumption and so on. To overcome these issues, FFSIFT-MLPC Technique is proposed.

The contribution of FFSIFT-MLPC Technique is summarized as follows,

- To improve the hyperspectral image classification accuracy, FFSIFT-MLPC technique is introduced with preprocessing, feature extraction and classification. The frost filtering technique is applied for removing the noise from the spectral band by changing the center pixel value of the window with the sum of the weighted average of the neighboring pixel. This helps to minimize the mean square error and improve the PSNR.

- To minimize the false positive rate, sigmoid activation function is used in the multilayer perceptron at the output layer. The activation function provides the binary outcomes by matching the extracted features with testing feature vectors through the Euclidean distance measure.

- To minimize the classification time, FFSIFT-MLPC technique extracts the feature using Gaussian distributive Scale-Invariant transformation. With the extracted features, the accurate classification is performed with minimum time.

**1.2    Structure of the paper**

The rest of this paper is ordered into different sections as follows. In Section 2, the proposed FFSIFT-MLPC Technique is described based on preprocessing and feature extraction. Section 3 shows the experimental results on hyperspectral images with different parameters. The experiments results are discussed in section 4 to demonstrate the performance of the proposed method. Finally, Section 5 provides the concluding remarks.

**2.    Methodology**

The proposed FFSIFT-MLPC Technique is introduced for classifying the hyperspectral images with higher accuracy. Hyperspectral images comprise a large amount of information to identify and classify the spectrally similar materials. Hyperspectral image (HSI) comprises hundreds of continuous narrow spectral bands that visible only in the infrared spectrum.  A multilayer perceptron is a machine learning technique which provides the binary classification results based on the set of the feature vector.  The FFSIFT-MLPC technique includes the three major processes namely preprocessing, feature extraction and classification. Initially, the HIS image acquisition is carried out from the database. The multilayer perceptron is the feed-forward artificial neural network of three or more layers i.e. one input layer and an output layer with one or more hidden layers of nonlinearly-activating nodes. The HIS image is divided into a number of spectral bands and it is given as input in the input layer.

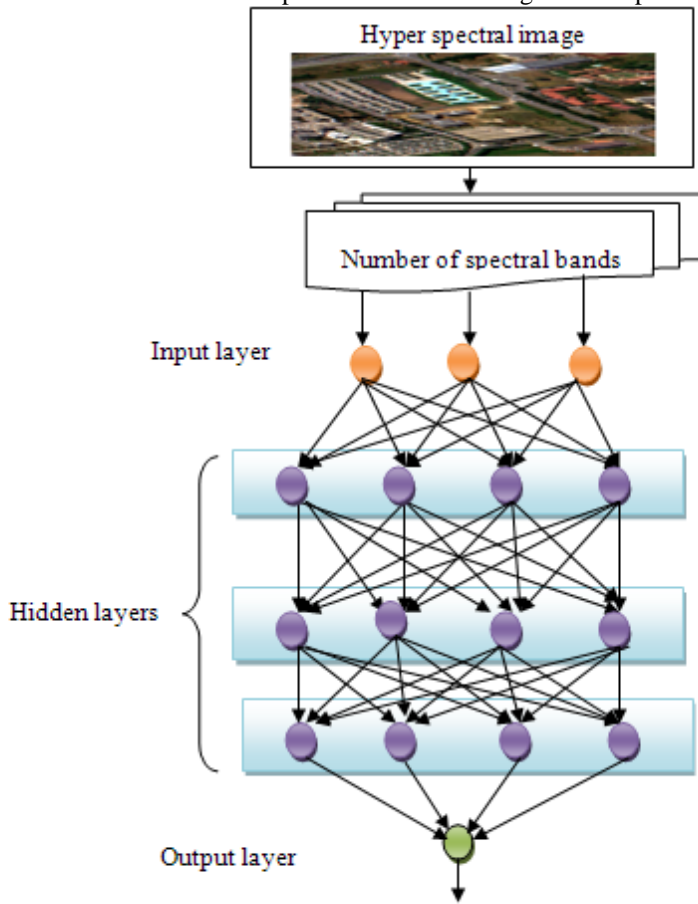

Figure 1 structure of the multilayer perceptron

Then the input is fed into the first hidden layer. In the hidden layer 1, the input images are preprocessed using frost filtering technique to remove the noise artifacts. In hidden layer 2, the feature extraction process is carried out using Gaussian distributive Scale-invariant feature transform. Then the extracted features are fed into the output layer for analyzing the extracted features with the testing features to classify the images. Figure 1 depicts a structural diagram of multilayer perceptron for HSI classifications. In figure 1, the neurons like the nodes in the one layer are fully connected with another layer to form a network. In a feed-forward neural network, the input moves one direction and it has no "fed back" into the input. As shown in figure 1, the output of one node to the input of another node is connected by an arrow symbol. The input layer receives the number of spectral bands into the network at a time '$t$' is denoted as '$x(t)$'. The output of the hidden layer and the output layer is denoted as '$h(t)$' and '$y(t)$' respectively. The node in one layer connects to another layer through dynamic weights $\tau_0, \tau_1$. The detailed explanation of the above said processes are described in the following subsections.

## 2.1. Frost filter based preprocessing

The preprocessing is done to improve the image quality by removing the noise from the spectral bands for object classification with minimum time and higher accuracy. The preprocessing is a process where a corrupt/noise in a spectral band is removed since it corrupted from motion blur, noise, and misfocused. This may cause an object classification in an inefficient manner. The proposed FFSIFT-MLPC technique uses the frost filter to remove the noise from an image.

Frost filter work based on the coefficient of variation and the window-based approach. The neighboring pixel patterns are called a window. The frost filter is the exponentially-weighted averaging filter. Within the window size of n*n, the center pixel value is changed by the weighted sum of values of the neighborhood pixels in the window. The example of 3*3 windows is shown in figure 2

| $p_0$ | $p_1$ | $p_2$ |
|---|---|---|
| $p_3$ | $p_{ij}$ | $p_5$ |
| $p_6$ | $p_7$ | $p_8$ |

Figure 2 3x3 window

Figure 2 shows the 3x3 windows where the pixels $p_0, p_1, p_2, \ldots \ldots p_n$ are arranged in the row and columns. As shown in figure 2, the center pixel $p_{ij}$ is replaced with the weighted sum of values of the neighborhood pixels. Frost filter based denoising is given below,

$$d_n = \sum_{nXn} \alpha\, \beta\, w \quad (1)$$
$$w = \exp(-\beta\, |T|) \quad (2)$$
$$\beta = \left(4 * \frac{1}{n\, D^2}\right)\left(\frac{D^2}{\mu^2}\right) \quad (3)$$

Where, $d_n$ is the image denoising, $n$ is the window size, $\alpha$ denotes a normalized constant, $\beta$ is the coefficient of variation which is defined as the ratio of local standard deviation ($D$) to the local mean ($\mu$) of the corrupted image, $w$ is the weighting factor, $|T| = (x - x_0) + (y - y_0)$ indicates the grid coordinates of the centre of the window and the pixel $p_{ij}$. The noisy pixel value is replaced by the weighted average of the all the pixels in the filter window. In this way, the noises in the input images are removed and enhanced the quality of images. The output of the first hidden layer is fed into the second hidden layer.

## 2.2. Gaussian distributive Scale-invariant transformation based feature extraction

In the second hidden layer, the feature extraction is performed using Gaussian distributive Scale-invariant transformation (GDSIT) for minimizing the complexity in the classification. The GDSIT is generally used to extract local features such as texture, color and object features of hyperspectral bands. By applying the GDSIT, initially construct the Difference of Gaussians that occurred at many scales. Therefore, the transformation is applied in a two-dimension,

$$d(x,y) = c(x,y,s\sigma) - c(x,y,\sigma) \quad (4)$$
$$c(x,y,\sigma) = g(x,y,\sigma) * i(x,y) \quad (5)$$
$$g(x,y,\sigma) = \frac{1}{\sigma\sqrt{2\pi}} \exp - \left(\frac{1}{2} * \frac{x^2+y^2}{\sigma^2}\right) \quad (6)$$

Where, $d(x,y)$ is the Difference of Gaussians, $c(x,y,s\sigma)$ represents the convolution of the original band $i(x,y)$ with Gaussian blur $g(x,y,\sigma)$, (x, y) represents the horizontal and vertical axis, $\sigma$ is the standard deviation of the Gaussian distribution, $s\sigma$ is the scale. Then the key points at the scale space extreme are obtained in the difference of Gaussian function. For each key point, orientation and gradient magnitude is assigned with the pixel difference as follows,

$$M = \sqrt{\left(c(x+1,y,) - c(x-1,y,)\right)^2 + \left(c(x,y+1) - c(x,y-1)\right)^2} \quad (7)$$

$$\emptyset = \tan^{-1}\left(\frac{c(x,y+1) - c(x,y-1)}{c(x+1,y,) - c(x-1,y,)}\right) \quad (8)$$

Where, $M$ denotes a magnitude, $\emptyset$ represents the orientation. The magnitude and direction estimation are done for each pixel in neighboring region around the keypoint. Therefore, the Gaussian distributive Scale-invariant transformation (GDSIT) based features are stored in the database for further processing.

### 2.3. Euclidean distance Measure

In the third hidden layer, extracted features vector are matched with the testing features vector of the hyperspectral image. The matching is done with the help of the Euclidean distance measure which is mathematically calculated as follows

$$l(f_e, f_t) = \sqrt{(f_e - f_t)^2} \quad (9)$$

Where, $l(f_e, f_t)$ distance between the extracted feature vector '$f_e$' and testing feature vector $f_t$. Then the distance between the features are fed into the output layer.

The output of hidden layer at the time '$t$' is expressed as follows,

$$h(t) = (\tau_o * x(t) + \tau h(t-1)) \quad (10)$$

From (10), $h(t)$ denotes an output of the hidden layer at the time stamp '$t$' and $h(t-1)$ denotes an output of the previous hidden layer, $x(t)$ denotes the input (i.e. spectral band), $\tau_o$ represents a weight between input and hidden layer, $\tau$ denotes a weights of the hidden layers at adjacent time stamps.

### 2.4 Image classification based on feature matching

At the output layer, the classification is done with the help of the matching the features using activation function. In multilayer perceptron, the activation function is used to define the output for a given input. The output of the multilayer perceptron is given below,

$$y(t) = A_f * \{h(t) * \tau_1\} \quad (11)$$

From (11), $y(t)$ represents an output at time '$t$', $A_f$ denotes the sigmoid activation function used in the final units, $\tau_1$ represents a weight between the hidden and output layer. $h(t)$ is the output of the hidden layer, The sigmoid activation function is mathematically formulated as follows,

$$A_f = (1 + \exp(-l))^{-1} \quad (12)$$

From (12), $A_f$ denotes a sigmoid activation function, $l$ denotes a distance between the extracted feature and testing features. The sigmoid activation function provides the two binary outcomes such as '0' and '1'. If the distance between the two features is minimal, the activation function returns '1'. Otherwise, it returns '0'. If the activation function provides '0', then the extracted feature in the spectral band is not matched with the testing feature vector. If the activation function provides '1', then the two feature vectors are exactly matched. Based on the matching results, the classified outputs are resulted in terms of spectral bands (i.e., different colors).

After the classification, the error rate is calculated in order to minimize the incorrect classification. The mean squared error is calculated as the average squared difference between the estimated results and the actual results. The error rate is computed using the following mathematical equations,

$$Err = (y_a(t) - y(t))^2 \quad (13)$$

Where $Err$ represents an Error, $y_a(t)$ denotes an actual output, $y(t)$ denotes an estimated output at the output layer of the classifier. Therefore, the classification process is repeated until the error is minimized for accurately classifying the HIS images with minimum false positive rate.

```
Input: Hyperspectral image
Output: Improve classification accuracy
Begin
    1.  for each image i
    2.     Divide the number of spectral bands b₁, b₂, b₃, .... bₙ
    3.     Given the spectral bands b₁, b₂, b₃, .... bₙ into input layer at time a 't' i.e. x (t)
    4.        for each spectral bands bᵢ
    5.           Filtering the noise in the first hidden layer
    6.              Extract the features 'fₑ' at second hidden layers h(t)
    7.              Measure the distance between the features l (fₑ, fₜ)
    8.              Obtain results at the output layer y(t)
    9.           If (Aᵣ = 1) then
    10.             fₑ are correctly matched with fₜ
    11.                Classify the images with different colors
    12.          else
    13.             fₑ are not matched with fₜ
    14.          end if
    15.          Compute training error Err
    16.          Update the weights between the layers
    17.          The process is iterated until find minimum classification error
    18.     end for
    19. end for
End
```

Algorithm 1 Frost Filtered Scale-Invariant Feature Transformation based MultiLayer Perceptron Classification

The algorithmic process of proposed FFSIFT-MLPC technique is clearly described in the step by step process. HSI is considered from the database. After that, HSI is partitioned into number of spectral bands. The bands are given as input for classification. The noise artifacts are removed to enhance the image contrast for accurate classification. Followed by, the features in the preprocessed bands are extracted in the second hidden layer. At the third layer, the Euclidean distance between the extracted feature vector ($f_e$) and testing feature vector is measured. At the output layer, feature matching process is carried out using distance value with the help of activation function. The sigmoid activation function is effectively matching the features and classified the images. Finally, the mean square error is computed based on the obtained results and actual output for each result. This process gets repeated until the proposed technique finds the minimum classification error. As a result, the proposed technique FFSIFT-MLPC technique improves the classification accuracy and minimizes the error rate.

3. **Experimental Setup and Parameter Settings**

The proposed FFSIFT-MLPC Technique and existing methods namely HCKBoost [1] and GF-SVM-EPF [2] are simulated using MATLAB. University of Pavia dataset [21] is employed as urban HSI for classifying the different scenes. The images are collected and divided into the spectral bands for classification. The different scenes are collected using ROSIS sensor using flight movement over Pavia University, northern Italy. There are 103 spectral bands are available in the dataset. For conducting the experimental, there are 100 bands are taken as input. Result analysis is performed with metrics namely,

- PSNR
- Classification accuracy
- False-positive rate
- Classification time

## 4. Results and discussions

In this section, the performance results of proposed FFSIFT-MLPC technique and existing methods namely HCKBoost [1] and GF-SVM-EPF [2] are discussed with the certain parameters such as PSNR, classification accuracy, false-positive rate, and classification time with respect to a number of spectral bands with the help of either tables or graphical representation.

### 4.1 Impact of peak signal to noise

PSNR is measured with mean square error based on squared difference between size of preprocessed band and original band with noise. The formula for mathematically calculating the mean square error and PSNR as given below,

$$Err_S = (b_i - b_p)^2 \quad (14)$$
$$PSNR = 10 * \log_{10}\left(\frac{r^2}{Err_S}\right) \quad (15)$$

Where, $Err_S$ denotes a mean square error, $b_i$ represents original band size and $b_p$ is a preprocessed band size, $r$ denotes a Maximum possible pixel value (i.e. 255). PSNR is calculated in decibel (dB).

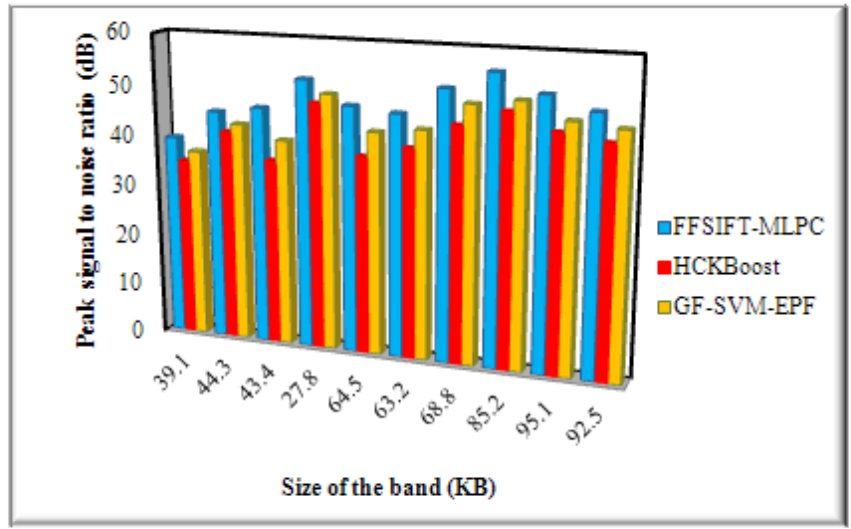

Figure 3 Performance results of Peak signal to noise ratio

Figure 3 depicts the experimental analysis of PSNR versus the size of the spectral band. From figure 3, the PSNR is found to be improved using FFSIFT-MLPC technique. This is because the utilization of frost filtering technique. The proposed FFSIFT-MLPC technique removes the noise artifacts and enhances the image contrast for accurate classification. The proposed filtering technique removes the noisy pixels from the spectral bands by replacing the center pixels with the weighted average of neighboring pixels and hence the image quality gets improved. The ten various results of PSNR of FFSIFT-MLPC technique is compared to the existing results. The average of ten results improves the peak signals to noise ratio by 16% and 8% as compared to HCKBoost [1] and GF-SVM-EPF [2].

### 4.2 Impact of classification accuracy

Classification accuracy is measured as the ratios of number of spectral bands are correctly classified to total number of spectral bands. The classification accuracy is calculated using below mathematical formula,

$$Cl_{Acc} = \left(\frac{Number\ of\ bands\ correclty\ classified}{N_{sp}}\right) * 100 \quad (16)$$

Where, $Cl_{Acc}$ represents the classification accuracy, '$N_{sp}$' denotes a number of spectral bands. Classification accuracy is measured in percentage (%).

Table 1 Classification accuracy versus number of spectral bands

| Number of spectral bands | Classification accuracy (%) | | |
|---|---|---|---|
| | FFSIFT-MLPC | HCKBoost | GF-SVM-EPF |
| 10 | 80 | 70 | 60 |
| 20 | 85 | 75 | 70 |
| 30 | 87 | 80 | 73 |
| 40 | 88 | 83 | 75 |
| 50 | 94 | 86 | 80 |
| 60 | 95 | 87 | 83 |
| 70 | 93 | 89 | 84 |
| 80 | 96 | 90 | 86 |
| 90 | 94 | 87 | 83 |
| 100 | 92 | 84 | 80 |

Table 1 illustrates the classification accuracy with number of spectral bands. There are 10 various results are obtained for various spectral bands. For the experimental consideration, the spectral bands are taken as input in the range from 10 to 100. The reported results show that the performance of classification accuracy using FFSIFT-MLPC technique is enhanced than the existing classification techniques. This is evidently proved using mathematical calculation. Let us consider the number of spectral bands is 10 for conducting the experiments. The FFSIFT-MLPC technique classified 8 spectral bands and the accuracy is 80% whereas the HCKBoost [1] and GF-SVM-EPF [2] accurately classifies 7 and 6 spectral bands and the accuracy are 70% and 60% respectively. Similarly, the different classification results are reported as shown in table 1. The above discussion clears that, the classification accuracy is found to be improved using FFSIFT-MLPC technique. The reason behind the feature matching is done at the output layer. The Gaussian distributive scale-invariant transform extracts the features from the bands and the Euclidean distance is used to measure the distance between the extracted features and testing features. Based on the distance measure, the feature matching is carried out and classified the results at the output layer. Therefore, classification accuracy based on the multilayer perceptron is found to be higher when compared to the state-of-the-art methods. The comparison results show that the classification accuracy is found to be improved by 9% when compared to HCKBoost [1] and 18% when compared to GF-SVM-EPF [2].

### 4.3 Impact of false-positive rate

False-positive rate is measured as the ratios of number of spectral bands are incorrectly classified to total number of spectral bands. It is calculated in percentage (%) and expressed as below,

$$F_{pr} = \left(\frac{Number\ of\ bands\ incorreclty\ classified}{N_{sp}}\right) * 100 \quad (17)$$

Where $F_{pr}$ represents the false positive rate, '$N_{sp}$' is a number of spectral bands.

In order to calculate the false positive rate, there are 10 spectral bands are considered as input. Among the ten bands, the 2 bands are incorrectly classified using FFSIFT-MLPC technique and 3, 4 bands are incorrectly classified using HCKBoost [1] and to GF-SVM-EPF [2]. With this, the percentage values of false-positive rate are 20%, 30%, 40% respectively. There are ten various results of false-positive rate are illustrated in figure 4.

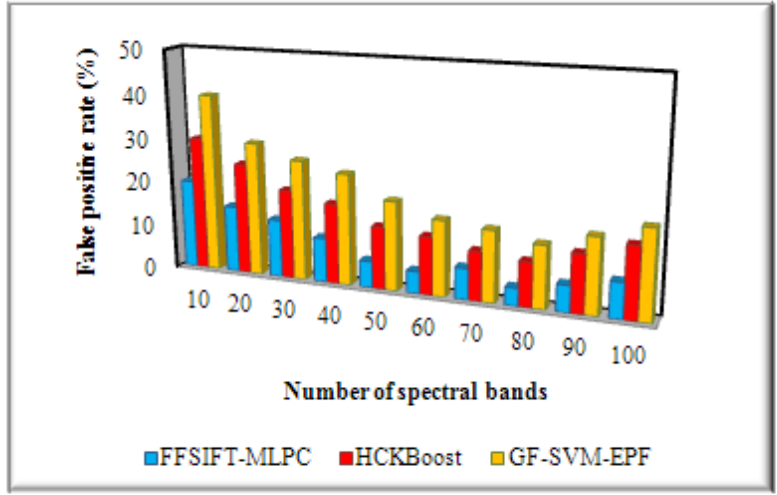

Figure 4 Performance results of false-positive rate

False positive rate of HSI classification is illustrated in figure 4 with number of spectral bands. The above graphical results show that the comparative analysis of three classification techniques FFSIFT-MLPC, HCKBoost [1] and GF-SVM-EPF [2]. As shown in the graph, the false positive rate of three methods is represented in the three different colors. The obtained results evidently prove that the FFSIFT-MLPC technique minimizes the false positive rate of hyperspectral image classification. This is owing to the application of the sigmoid activation function used in the multilayer perceptron. The activation function provides binary outcomes by matching the feature vectors. When the extracted features and the testing features vector distance are minimal, the activation function accurately performs the classification. This helps to minimize the misclassification resulting in reduces the false positive rate. The observed results of the three classification methods are compared. The output results minimize the false positive rate of classification by 47 % using FFSIFT-MLPC technique as compared to existing HCKBoost [1] and 60% compared to GF-SVM-EPF [2] respectively.

### 4.4. Performance analysis of classification time

Classification time is the amount of time consumed by an algorithm to classify the spectral bands. The overall classification time is computed as follows,

$$C_{time} = N_{sp} * T \ (classifying \ one \ spectral \ band \ ) \quad (18)$$

Where $C_{time}$ represents the classification time, $N_{sp}$ is the number of input spectral bands, $T$ denotes a time for classifying one band. The classification time is measured in the unit of milliseconds (ms).

Table 2 Classification time

| Number of spectral bands | Classification time (ms) | | |
|---|---|---|---|
| | FFSIFT-MLPC | HCKBoost | GF-SVM-EPF |
| 10 | 15 | 17 | 19 |
| 20 | 18 | 22 | 26 |
| 30 | 24 | 27 | 30 |
| 40 | 28 | 34 | 38 |
| 50 | 30 | 36 | 40 |
| 60 | 34 | 41 | 44 |
| 70 | 37 | 43 | 48 |
| 80 | 42 | 45 | 50 |
| 90 | 43 | 47 | 52 |
| 100 | 49 | 53 | 55 |

Table 2 describes the performance result of classification time with a number of spectral bands taken in the range from 10 to 100. By increasing the number of spectral bands, the classification time also differs due to the different number of input was taken. The output results clear that the FFSIFT-MLPC technique minimizes the classification time than the state-of-the-art methods. The reason behind the FFSIFT-MLPC technique performed preprocessing as well as feature extraction in the hidden layers. The preprocessing helps to enhance the quality of images in the spectral bands by removing the noisy pixels. Followed by, the transformation technique extracts the different features such as texture, color and object features of hyperspectral bands. With the extracted features, the FFSIFT-MLPC technique performs the accurate classification based on feature matching with minimum time. When taking 10 spectral bands from the Dataset to conduct experimental work, proposed FFSIFT-MLPC technique consumes $15ms$ of time for classification whereas existing works utilizes $17ms$ and $19ms$ respectively. Accordingly, the average of ten various results of classification time using proposed FFSIFT-MLPC technique is minimized by 13% and 22% as compared to other works HCKBoost [1] and GF-SVM-EPF [2].

The above-discussed result of the different parameters confirms that the FFSIFT-MLPC technique accurately classifying the hyperspectral images with minimum time and false-positive rate.

## 5. Conclusion

A novel methodology FFSIFT-MLPC technique is developed with the aim of improving the classification accuracy of the hyperspectral images with minimum time. The multilayer based classification is considered as a computationally efficient algorithm. The multiple layers are used in the FFSIFT-MLPC technique for effectively performs the preprocessing by applying the frost filtering technique. Followed by, the multiple features are extracted from the preprocessed bands. Finally, the Euclidian distance between the feature vectors is calculated to accurately classify the images through the activation function at the output layer. The experiment is carried out under the various parameters such as the PSNR, classification accuracy, false positive rate and classification time. The observed results show that the proposed technique improves the hyperspectral image classification accuracy than the state-of-art- methods.